\documentclass{uai2026} 
                        
\usepackage[american]{babel}
\usepackage{algorithm}
\usepackage{algorithmic}
\usepackage{amsfonts}
\usepackage{natbib} 
    \renewcommand{\bibsection}{\subsubsection*{References}}
\usepackage{amsthm}
\usepackage{xcolor}
\usepackage{array}
\usepackage{blkarray, bigstrut,array}
\usepackage{multirow}
\usepackage{color, colortbl}
\definecolor{darkred}{rgb}{0.55, 0.0, 0.0}
\usepackage{amsmath}
\usepackage{subcaption}
\usepackage{graphicx}    
\usepackage[table,xcdraw]{xcolor}
\newtheorem{theorem}{Theorem}[section]
\newtheorem{definition}{Definition}[section]
\usepackage{mathtools} 
\usepackage{booktabs} 
\usepackage{tikz} 

\usepackage{color-edits}
\addauthor{mt}{darkred}
\addauthor{mc}{blue}
\addauthor{mb}{magenta}

\title{Identifiable Multimodal Causal Representation Learning under Partial Latent Sharing}

\author[1]{\href{mailto:<manal.benhamza@centralesupelec.fr>?Subject=Your UAI 2026 paper}{Manal Benhamza}{}}
\author[2]{Marianne Clausel}
\author[1]{Myriam Tami}
\affil[1]{%
    Paris-Saclay University\\
    CentraleSupélec\\
    MICS Lab\\
    France
}
\affil[2]{%
    Lorraine University\\
    CRAN\\
    France
}
  
\begin{document}
\maketitle
\begin{abstract}
  Causal representation learning (CRL) seeks to uncover meaningful latent variables and their corresponding causal structure from high-dimensional observational data. Although its significance,  CRL’s identifiability remains a crucial property, as it ensures the recovery of the mechanisms behind the data generation process, and hence the interpretability and robustness of the representation. Proving identifiability in CRL is intrinsically difficult, and we address in this work an even more challenging setting: multimodality. We consider multimodal observed data with a latent partially shared structure. Each modality is generated, through non linear mixing functions, from a specific subset of causal latent variables. Under flexible assumptions and without imposing any parametric distribution on the latent variables, we establish \textit{component-wise identifiability guarantees} for the causal latent representation. Our identifiability results, furthermore, apply to the undercomplete scenario where we have, for each modality, more observed than latent variables. To instantiate our theoretical analysis, we introduce a Wasserstein-based module to recover the partially shared latent structure. Due to its differentiability, the latter can be easily integrated into all types of architecture, only requiring minimal changes. Extensive experiments on synthetic and realistic datasets validate the superiority of our approach over SOTA methods.
\end{abstract}

\section{Introduction}
\label{intro}
Causal Representation Learning (CRL) aims to uncover high-level meaningful representations and their corresponding causal structure from low-level observational data. It has attracted growing importance across a wide variety of fields \citep{fromiden}. For example, CRL is leveraged in medical applications to avoid learning spurious correlations, thereby promoting more robust representations \citep{precisionmed}. It is also used in biological sciences with multimodal sequencing data to learn the causal latent variables and their related causal relationships, hence providing more transparency to complex biological systems \citep{squires}. More generally, it is worth noting that causal representations enable multiple aspects of trustworthy AI by achieving fairness \citep{towards,fairness}, interpretability \citep{interpretability} and generalizing to out-of-distribution samples \citep{luetal, Sunetal,beery,generalizability}. However, despite these promising advantages, the practical use of CRL in real-world scenarios remains constrained by strong identifiability assumptions. Establishing identifiability guarantees is indeed crucial: in their absence, two independently learned representations may achieve the same learning objective while being structurally different. As a result, the learned representation may fail to capture the true underlying data-generating process, thereby undermining interpretability.

\begin{figure}[htbp]
    \centering
    \hspace{-0.4cm} \includegraphics[scale=0.4]{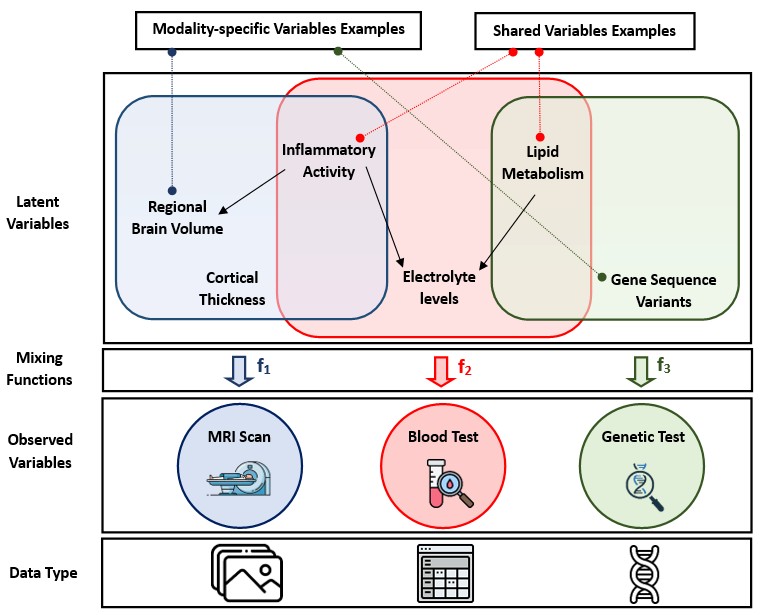}
    \caption{Multimodal MRI Dataset encompassing both modality-specific and overlapping shared latent variables. The black directed arrows represent the causal relationships among the causal latent variables. Here, for example, the Inflammatory Activity is a shared variable between the MRI scan and blood test modalities.}
    \label{mri_example}
\end{figure}

In this work, we consider a multimodal setting where the observed variables are functions of subsets of causal latent variables. This setting naturally appears in many practical scenarios. For example, as illustrated in Fig. \ref{mri_example}, in medical datasets, an MRI can show structural changes in the brain, a blood test can reveal inflammatory responses, and a genetic test can indicate predispositions to certain diseases. Each modality reflects only a subset of the underlying variables, some of which may overlap with other modalities, while others remain specific to a single modality.

Recent works have addressed the identifiability issue by assuming a constraining parametric structure on the distribution of latent variables \citep{morio2024,gresele, morio2023}. Other approaches are based on auxiliary signals such as supervision \citep{causalvae, dear, scmvae}, interventions on latent variables \citep{learlincau} or temporal dependence \citep{sparsity}, which are not always accessible. The multimodal setting has attracted significant interest, as it enables the extraction and integration of complementary information across multiple modalities \citep{complement}. In this context, the focus is mainly on identifying the subspace of shared variables between the different modalities \citep{multiview, morio2024, mcl}. 
The sharing pattern can either be total, with all modalities depending on a common set of latent variables, or partial, where only subsets of modalities share latent variables, as shown in Fig. \ref{mri_example}. Identifiability guarantees for these multimodal paradigms are limited. 
Moreover, most existing approaches assume that the modality mixing functions are bijective \citep{multiview, mcl, biome, vonkug}. This is often unrealistic in real-world scenarios, such as image data, where the dimensionality of the observations can be much higher than that of the latent ones. Consequently, the identifiability assumptions underlying prior work remain restrictive.

 We, thus, establish in this work, identifiability guarantees for the causal representation in a multimodal with a partially shared latent structure setting. More specifically, we prove under structural causal sparsity constraints that both the shared and modality specific variables can be identified, up to component-wise smooth bijections. Our work, thus, generalizes the block-wise identification results in \citep{multiview,mcl} and ensures interpretability of the learned representations. As opposite to other work, our guarantees also hold in undercomplete cases where there are, in each modality, more observed variables than latent ones and we do not require access to supervision nor to interventions. Our theoretical conditions are hence weaker than other restrictive works. Furthermore, to show the efficiency of our theoretical conditions, we implement them in a generative framework to estimate the causal latent variables and their causal relationships. We also introduce a Wasserstein distance–based, architecture-agnostic pluggable module to recover the partially shared latent structure. Extensive experiments on both synthetic and realistic datasets validate the superiority of our approach over SOTA methods.
\section{Related Work}
\paragraph{\textbf{Multimodal Identifiable CRL:}} As emphasized in Sec. \ref{intro}, identifiability is a central concern in CRL.
We focus on the \textit{multimodal} CRL setting, which has recently attracted increasing attention \citep{biome,morio2024}.
\begin{enumerate}[label=(\alph*)]
\item \textbf{Non Overlapping Latent Structure:} \citet{biome} prove the component-wise identifiability of latent variables from each modality by enforcing sparsity over the causal structure. They require, however, the mixing functions to be bijective. \citet{morio2024} consider a setup where the grouping of the observed variables is known a priori. Their identifiability results require the latent variables to follow an exponential family distribution, which may limit applicability in practice. \citet{biome,morio2024} guarantees are established for non-overlapping latent structures. While their frameworks are argued to extend to settings with shared latent variables, such extensions are not explicitly developed, and handling partially shared structures remains an open challenge in practice.
\item \textbf{Overlapping Latent Structure:} In the multimodal setting with shared latent variables, the focus is mainly on identifying the shared variables. \citet{sturma} focus on the identifiability of the joint distribution and the shared causal graph in a setup where the modalities potentially share a representation. Nonetheless, their guarantees are limited because they only hold for linear mixing functions. \citet{vonkug,mcl} give sufficient conditions for the identifiability of the shared latent variables in a pair of modalities setup. Both approaches assume that the second modality is generated as an augmentation of the first, through perturbations of a subset of modality-specific variables, commonly denoted style variables. As a result, this setting may not reflect more general multimodal generative mechanisms. In \citet{vonkug}, the mixing functions are further assumed to be identical across modalities, whereas \citet{mcl} relax this assumption and allow for distinct mixing processes. Nevertheless, the resulting guarantees remain limited, as they only ensure subspace-level identifiability and rely on strong assumptions, in particular that the modality mixing functions are diffeomorphisms.
\citet{multiview} presents a multi-view setting, which is a particular case of the multimodal. The identifiability results are only block-wise established for the subset of shared variables. Furthermore, both the theoretical analysis and empirical framework are contrastive learning based, hence, requiring carefully constructed positive and negative data pairs.
\end{enumerate}
Our work extends previous approaches to a more general multimodal setting with partially shared latent structure and, importantly, different injective non linear mixing functions across modalities. In particular, the modalities are not necessarily augmented versions of each other. Within this general framework, we establish the component-wise identifiability of both shared and modality-specific variables.

\paragraph{\textbf{Partially Shared Pattern Learning:}} Only few works address the partially shared pattern learning problem. Most of them are contrastive learning based \citep{vonkug,mcl,multiview}. \citet{partialregression} propose to use a trainable classifier with a downstream task to identify the latent shared representations between modalities. We introduce a novel Wasserstein-based differentiable pluggable module which computes, through a transport optimization problem \citep{transpop}, the shared structure permutation matrix introduced in Sec. \ref{framework}. By unveiling the shared latent variables between the different modalities, this matrix promotes a better cross-modal alignment. Thanks to our proposed module, we eliminate the need for additional training, supplementary data, or auxiliary supervision, while remaining easily compatible with a wide range of encoder–decoder architectures.

\section{Model Definition}
We formulate our multimodal data generation process as a latent variable model. Let $\mathbf{z}=(\mathbf{z}_1,…,\mathbf{z}_J)$ be a causally related continuous random variable taking values in $\mathcal{Z} = \mathcal{Z}_1 \times ... \times \mathcal{Z}_J$ and collectively capturing all the necessary information to generate the entangled observed variables $\mathsf{x}=(\mathsf{x}^{(1)},….,\mathsf{x}^{(M)})$, where $\mathsf{x}^{(m)}$ denotes the $m^{th}$ modality with dimension $d_m$. We consider that each modality $m$ only depends on a subset of latent variables $\mathbf{z}_{\mathcal{A}_m}=\{\mathbf{z}_j,j \in {\mathcal{A}_m}\}$. While each observed variable only belongs to one of the $M$ observed modalities, there might be overlapping between the subsets of latent variables $\{\mathbf{z}_{\mathcal{A}_m}\}_{m=1}^M$. 


\noindent{\bf Generating process $Gen(\mathbf{z},\mathbf{\epsilon},g_1,\cdots,g_J,f_1,\cdots,f_m)$.} We model the data-generating process as follows:
\begin{align}
& \mathbf{z}_j = g_j( \text{Pa}(\mathbf{z}_j),\mathbf{\epsilon}_j) \quad \forall j \in \{1,\cdots,J\}\label{causalmechanism}\\
& \mathsf{x}^{(m)}=f_m(\mathbf{z}_{\mathcal{A}_m}) 
\label{generationproc}
    \end{align}

We assume that we have causal relationships between the latent variables, as specified in Eq. \ref{causalmechanism}. $\text{Pa}(\mathbf{z}_j)$ denotes the causal parents of the variable $\mathbf{z}_j$, $g_j$ represents the corresponding causal mechanism, and $\epsilon_j$ are the exogenous noise variables, which are assumed to be mutually independent. In Eq. \ref{generationproc} we generate, for each modality $m$, observed variables $\mathsf{x}^{(m)}$ from latent ones $\mathbf{z}_{\mathcal{A}_m}$. The modality specific mixing functions in Eq. \ref{generationproc}, denoted by $f_m$ are assumed to be smooth and injective, hence allowing for less latent than observed variables in each modality $m$, i.e., $ \forall m \in \{1,...,M\} \quad |\mathcal{A}_m| \leq d_m$. We further assume that the different observed modalities are obtained through different mixing functions $\forall m,n \in \{1,...,M\} \quad f_m \neq f_n$. 

\noindent{\bf Example 1: Multimodal dataset with a partially shared causal latent structure. }
\begin{figure}[ht!]
    \centering
    \includegraphics[scale=0.6]{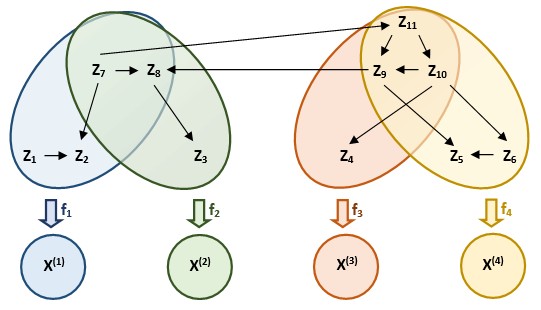}
    \caption{The variables $\mathbf{z}_7$ and $\mathbf{z}_8$ are shared by the first and second modalities, whereas $\mathbf{z}_3$ and $\mathbf{z}_4$ are modality-specific.}
    \label{datasetup}
\end{figure}
Our objective is to prove component-wise identifiability results for the shared and modality specific latent representations under weak assumptions. More specifically, we establish under hypothesis on the data generating process, that we can recover from observed variables, for each modality $m$, the ground-truth shared and modality-specific  latent variables $\mathbf{z}_{\mathcal{A}_m}$ up to component-wise bijections. That is, we want an estimation $\hat{\mathbf{z}}_{\mathcal{A}_m}$ of the ground truth latent variables $\mathbf{z}_{\mathcal{A}_m}$ of the form $\forall \ \hat{\mathbf{z}}_j \in \hat{\mathbf{z}}_{\mathcal{A}_m} \quad \exists \hspace{0.1cm} h_j^{(m)},\sigma^{(m)} \hspace{0.1cm} \text{such that} \hspace{0.1cm} \hat{\mathbf{z}}_j=h_j^{(m)}((\mathbf{z}_{\mathcal{A}_m})_{\sigma^{(m)}(i)}) $. We want to emphasize that our proposed identifiability guarantees, ensure the estimated shared variables (resp. modality specific) depend only on their ground truth counterpart. Then, shared variables are disentangled from the modality specific ones. From a practical point of view, this implies efficient retrieval procedures of latent variables, relying on our Wasserstein-based module. Due to their invariance across modalities, these shared variables are efficient representations for downstream prediction tasks \citep{class1}.  We also aim to learn the identified causal latent representations and their related causal structure. Learning causal relationships on top of the latent representations enables a principled understanding of the underlying data-generating mechanisms. 

\section{Identifiability Theory}
\label{theory}
\textbf{Notations:} We denote $J_{f_m}$ (resp. $J_{\hat{f}_m}$, $J_{g}$, $J_{\hat g}$) the Jacobian of $f_m$ (resp. $\hat{f}_m$, $g:=(g_1,\cdots, g_J)$ and $\hat g=(\hat g_1,\cdots,\hat g_J)$). 
$dim(\cdot)$ denotes the dimensionality of a given set of variables, and $I(\cdot)$ the indices of the corresponding set. 
Superscripts and subscripts enclosed in parentheses $(m)$ refer to the $m^{\text{th}}$ modality. 
The $\mathbf{c}^{(m_1,\cdots,m_p)}$ are the common variables between the $p$ modalities $m_1,\cdots,m_p$ whereas $\mathbf{s}^{(m)}$ are the variables specific to modality $m$. We also denote $Sh(m)=\{l \in \{1,...,M\} \hspace{0.1cm} \text{s.t} \hspace{0.1cm} \mathbf{z}_{\mathcal{A}_l} \cap \mathbf{z}_{\mathcal{A}_m} \neq \emptyset \}$ the indices of modalities that share variables with the $m^{th}$ modality.

\subsection{Overview of the contributions} 
In this section, we establish component-wise identifiability guarantees for both modality-specific variables and variables shared across arbitrary subsets of modalities. Our approach aims to relax the commonly imposed strong assumption of bijectivity on the modality mixing functions $f_m$. In particular, it also applies to the undercomplete setting, where the mixing functions are assumed to be injective rather than diffeomorphic.
 Furthermore, we consider a general multimodal setting in which shared variables exhibit a partially shared structure, instead of being common to all modalities. This implies that the common latent variables $\mathbf{c}^{(m)}$ are shared only among subsets of modalities, rather than across the entire set of modalities. Our identifiability analysis and results rely on flexible conditions and do not require access to supervisory signals, interventional data, or assumptions on the parametric form of the latent distribution. 
Before introducing our theoretical guarantees and results, it is essential, to first recall the classical definitions of block and component-wise identifiability.
\begin{definition}[Block-wise Identifiability] 
    The true latent variables $\mathbf{z}$ are said to be block-identified, 
    if the estimated latent variables $\hat{\mathbf{z}}$ contain all and only information about the true latent variables $\mathbf{z}$, i.e. there exists a smooth invertible mapping $h:\mathbb{R}^{dim(Z)} \to \mathbb{R}^{dim(Z)}$ such that $\hat{\mathbf{z}}=h(\mathbf{z})$. 
\end{definition}
\begin{definition}[Component-wise Identifiability]
    A true latent variable $\mathbf{z}_i$ is said to be component-wise identified, 
    if a component of the estimated latent variables $\hat{\mathbf{z}}_j$ contains all and only information about $\mathbf{z}_i$, i.e., there exists a smooth invertible mapping $h_i:\mathbb{R}^{dim(Z_i)} \to \mathbb{R}^{dim(Z_j)}$, such that $\hat{\mathbf{z}}_j=h_i(\mathbf{z}_i)$.
\end{definition}
We can hence understand that block-wise identifiability ensures that the information is preserved within the subspace of estimated variables, whereas component-wise identifiability guarantees that each latent component is individually recoverable up to an invertible transformation.

\noindent{\bf Example 2 (Example 1, continued).} In Fig. \ref{datasetup}, the block wise identifiability of the shared variables $\mathbf{c}^{(1)}=\{\mathbf{z}_7,\mathbf{z}_8\}$ between modalities $1$ and $2$ implies the existence of an invertible function $h^{(1)}:\mathbb{R}^{d(\mathbf{c}^{(1)})} \to \mathbb{R}^{d(\mathbf{c}^{(1)})}$ s.t. $\hat{\mathbf{c}}^{(1)}=h^{(1)}(\mathbf{c}^{(1)})$. As for the component-wise identifiability of $\mathbf{z}_8$, an option with one dimensional variables would be to have an invertible function $h_7:\mathbb{R} \to \mathbb{R}$ such that $\hat{\mathbf{z}}_8=h_7(\mathbf{z}_7)$.


\paragraph{Theoretical contributions:} We present two main identifiability results : (1) the block-wise identifiability of the shared variables between each subset of modalities; (2) the component-wise identifiability of both specific $\mathbf{s}^{(m)}$ and common latent variables $\mathbf{c}^{(m)}$. 

\noindent{\bf Sketch of the proofs.} First, we establish for each modality $m$ that there exists an invertible function mapping the estimated latent variables $\hat{\mathbf{z}}_{\mathcal{A}_m}$ to their ground-truth counterparts $\mathbf{z}_{\mathcal{A}_m}$ under injective and distinct modality-mixing functions. Then we show that the estimated shared variables across a subset of modalities depend only on their ground-truth counterparts. We, thereafter, prove that under causal structural sparsity constraints as in \citep{biome,zhengzhang,sparsity}, both shared and specific latent variables in a modality $m$ can be uniquely identified. 

\subsection{Block identification}
We now establish, as previously outlined, the block-wise identifiability of the shared latent variables between each subset of modalities. 
\begin{theorem}[Block-wise Identifiability of modality shared variables]
\label{sharedblockwise}
Assuming that the observed data are obtained from the true latent variables (resp. estimated ones) through the ground truth data-generating process $Gen(\mathbf{z},\mathbf{\epsilon},g_1,\cdots,g_J,f_1,\cdots,f_M)$ (resp. estimating data generating process $Gen(\hat{\mathbf{z}},\hat{\mathbf{\epsilon}},\hat g_1,\cdots,\hat g_J,\hat f_1,\cdots,\hat f_M)$) defined in Eq.\ref{generationproc}, and given the assumptions below:
\begin{itemize}
        \item[$\bullet$] A$1$ (\textbf{Linear Independence}): The Jacobians of the true and estimated generating functions $J_{f_m}$ and $J_{\hat{f}_m}$ are full-column rank.
        \item[$\bullet$] A$2$ (\textbf{Properness}): Both $f_m$ and $\hat{f_m}$ are proper functions:
        \begin{equation}
\|\mathbf{z}\|, \|\hat{\mathbf{z}}\| \xrightarrow{} \infty \iff f_m(\mathbf{z}), \hat{f_m}(\hat{\mathbf{z}}) \xrightarrow{} \infty 
\end{equation}
\item[$\bullet$] A$3$ (\textbf{Conditional and Marginal Independence}):  Causal edges are allowed only from shared variables to specific variables, and not vice versa. Moreover, conditioned on $\mathbf{c}^{(m,n)}$ and $\mathbf{c}^{(k)}$ $\forall k$ s.t  $\hspace{0.1cm} Sh(m)\cap Sh(k)=\emptyset$, $\forall n \in Sh(m)$, $\mathbf{s}^{(m)}_i$ are independent from $\mathbf{s}^{(n)}_j$ $\forall \hspace{0.1cm} i,j \in I(\mathbf{s}^{(m)}) \times I(\mathbf{s}^{(n)})$. Moreover, we require that $\mathbf{c}^{(m)}_i$ and $\mathbf{c}^{(n)}_j$ are either marginally independent $\forall n \in Sh(m)$ and $\forall \hspace{0.1cm} i,j \in I(\mathbf{c}^{(m)}\setminus \mathbf{c}^{(m,n)}) \times I(\mathbf{c}^{(n)}\setminus \mathbf{c}^{(m,n)})$ or conditionally independent given $\mathbf{c}^{(k)}$ $\forall k$ s.t  $\hspace{0.1cm} Sh(m)\cap Sh(k)=\emptyset$.
    \end{itemize}
Then the shared variables for each subset of modalities are block-identified from observed variables. 
\end{theorem} 

\paragraph{Discussion on the assumptions:} The proof is on the Appendix A. Th. \ref{sharedblockwise} first assumption Th. \ref{sharedblockwise} A$1$ is common in \citep{condition}. It ensures, for a modality $m$, that each latent variable affects the observed variables independently of the others. This assumption is milder than requiring the true and estimated mixing functions to be diffeomorphic, as is commonly assumed in most identification theories \citep{biome,multiview,mcl,vonkug}. Properness Th. \ref{sharedblockwise} A$2$ ensures that as latent variables diverge, the corresponding observed variables also diverge, preventing any loss of information at extreme values. It guarantees that the mapping from latent to observed space is globally well-behaved, so that all latent variations are faithfully reflected in the observations and are, in principle, recoverable. As for the conditional and marginal independence Th. \ref{sharedblockwise} A$3$, it assumes there are no causal relationships between specific latent variables from two different modalities. The causal influence of each latent specific variable is, therefore, confined in its modality, whereas that of shared variables may span other modalities as shown in Fig. \ref{datasetup}.

\subsection{Component-wise Identifiability}
Following on Th. \ref{sharedblockwise} results, we establish the component-wise identifiability guarantees. Not to load notations, we use interchangeably here $I(z_{\mathcal{A}_m})$ and $(m)$.

\noindent{\bf Preliminary notations.} Set $\mathcal{T}:=\{T \in \mathbb{R}^{J \times J},\,\forall m,\, T_{(m),(m)}\mbox{ is invertible}\}$, $\mathcal{S}_m=\{P\in \mathbb{R}^{\mathcal{A}_m\times \mathcal{A}_m},\, \exists \sigma \mbox{ permutation of }\mathcal{A}_m,\, \forall j\neq \sigma(i),\, P_{i,j}=0 \}$ and $\mathcal{T}^{(-m)}=\{T\in \mathcal{T},\, T_{(m),(m)}\notin \mathcal{S}_m\}$. 

The following result also involves the ground truth and estimated latent causal structure encoded in the two matrices $J_{g}$ and $J_{\hat g}$, that can be identified as the adjacency matrices of the ground truth and estimated latent graphs $G$ and $\hat G$ respectively. 
\begin{theorem}(Component-wise Identifiability of modality-specific and shared variables)
\label{componentident}
    Given the same assumptions in Th. \ref{sharedblockwise}. Assume that in addition, the following conditions hold:
    \begin{itemize}
        \item [$\bullet$] B$1$ (\textbf{Non-overlap}): For each modality $m$, there is another modality $k$, such that there is no shared structure between the two modalities , i.e., $Sh(m) \cap Sh(k)=\emptyset$.
        \item [$\bullet$] B$2$ (\textbf{Mixing Density}): For any $T \in \mathcal{T}^{(-m)}$ and any $k$ such that i.e., $Sh(m) \cap Sh(k)=\emptyset$, then: 
        \begin{equation}
\hspace{-0.2cm}\displaystyle \sum_{\underset{\small Sh(m) \cap Sh(k)=\emptyset}{m,k}} \hspace{-0.5cm}\|\tilde{J}_{g}^{(T,m,k)}\|_0 >\hspace{-0.5cm} \displaystyle \sum_{\underset{Sh(m) \cap Sh(k)=\emptyset}{m,k}}\hspace{-0.5cm}\|[J_g]_{(m),(k)} \|_0
\end{equation}
with $\tilde{J}_{g}^{(T,m,k)}:=T_{(m),(m)} [J_g]_{(m),(k)} T_{(k),(k)}^{-1}$
        \item [$\bullet$] B$3$ (\textbf{Causal Structural Sparsity}): The estimated causal graph $\hat G$ is sparser than its true counterpart $G$ in the non sharing modalities, i.e.,
        \begin{equation}
           \displaystyle \sum_{\underset{Sh(m) \cap Sh(k)=\emptyset}{m,k}} \hspace{-0.3cm} \|[J_{\hat{g}}]_{(m),(k)}\|_0 \leq \displaystyle \sum_{\underset{Sh(m) \cap Sh(k)=\emptyset}{m,k}}\hspace{-0.3cm}\|  [J_g]_{(m),(k)}\|_0
        \end{equation}
Then the specific and shared latent variables are component-wise identified. 
    \end{itemize}
\end{theorem}
\begin{figure}[htbp]
    \centering
    \includegraphics[scale=0.5]{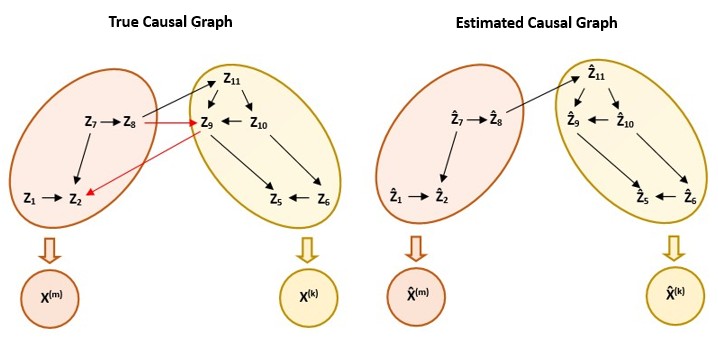}
    \caption{Th. \ref{componentident} causal structural sparsity condition illustration. The estimated causal graph is sparser than the true one. Additional edges are highlighted in red.}
    \label{identcondition}
\end{figure}
\textbf{Intuition:} 
In Theorem \ref{componentident}, Assumption B$2$ accounts for the additional causal edges that can appear in blocks $[J_g]_{(m),(k)}$ when identifiability breaks. It stipulates that if $T_{(m),(m)}$ is a non generalized permutation matrix, then there exists a group $k$ such that $[J_g]_{(m),(k)}$ is structured so that multiplication by $T_{(m),(m)}$ results in nontrivial linear mixing of rows and columns. Such mixing neccesarily reduces the sparsity of the block. An example is given on Appendix A.

\paragraph{Discussion on the assumptions:} See the proof of the Th. \ref{componentident} in Appendix A. The first assumption in Th. \ref{componentident} B$1$ requires the latent components to exhibit only partial sharing. Concretely, this means that no latent variable is shared across all the modalities. They are rather shared by a subset of modalities. This partial pattern ensures the sharing structure is distinct among the different modalities. Both the second Th. \ref{componentident} B$2$ and third Th. \ref{componentident} B$3$ assumptions, namely the mixing density and structural sparsity assumptions are an adaption of \citep{biome} causal component-wise identifiability conditions, to the partial overlapping setup. Under assumption B$2$, the blocks $[J_g]_{(m),(k)}$ may contain additional edges when the identifiability breaks. Assumption Th. \ref{componentident} B$3$ implies the absence of some inter-modal causal relationships in the estimated causal graph, as illustrated in Fig. \ref{componentident}. The latter sparsity constraint is very common in identifiaction works \citep{fumero,sparsity,xu} and is satisfied in many real-world scenarios \citep{case1, case2}. The second and third assumptions together induce a contradiction if there is an $m$ such that $[J_{\frac{\partial \hat{\mathbf{z}}}{\partial \mathbf{z}}}]_{(m),(m)}$ is not a generalized permutation matrix. In particular, Th. \ref{componentident}B$2$ and Th. \ref{componentident}B$3$, are only verified when there is a one to-one mapping between the latent variables in each modality. Combined, the assumptions of Th. \ref{componentident} prove the component-wise identifiability of shared and modality-specific latent variables.
\section{Learning Framework}
\label{framework}
\begin{figure*}[h!]
\centering
\begin{minipage}{0.35\textwidth}
    \centering
    \includegraphics[scale=0.44]{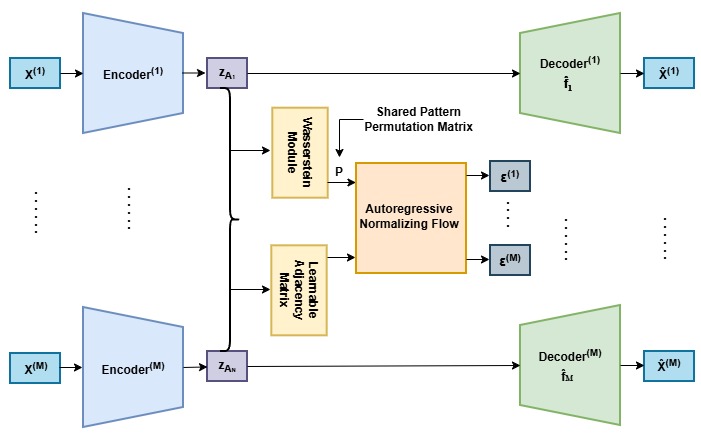}
\end{minipage}
\hfill
\begin{minipage}{0.35\textwidth}
    \par
    
    \caption{\centering \textbf{The proposed estimation framework.} It infers based on each modality's observed variables $\mathsf{x}^{(m)}$ the corresponding causal latent variables $\mathbf{z}_{\mathcal{A}_m}$. The latter are then passed through modality-specific decoders to reconstruct the obervations $\hat{\mathsf{x}}^{(m)}$. The model also determines, leveraging its Wasserstein-based pluggable differentiable module, the partially shared data pattern. The exogenous variables $\epsilon^{(m)}$  are estimated with an autoregressive normalizing flow and causal relationships are captured in a learnable adjacency matrix, as is standard in the literature \citep{causnormflows}.}
    \label{fig:example}
\end{minipage}
\end{figure*}
We propose a learning framework to corroborate our theoretical identification conditions. Our framework learns causal latent representations, using Th. \ref{componentident}. It also captures the partially shared structure by determining and thereafter aligning the shared elements between different modalities. Our model also infers the causal structure over the estimated latent variables. Our estimation framework is illustrated in Fig. \ref{framework}. The framework includes (i) an autoencoder structure; (ii) a Wasserstein module learning the latent shared structure; (iii) the normalizing flow learning the causal latent structure. This last step heavily relies on our identifiability results proven in Sec. \ref{theory}.
\subsection{Auto encoder structure}
\paragraph{Encoder:} Every modality's observed variables $\mathsf{x}^{(m)}$ are passed through a modality-specific encoder to estimate the modality's corresponding causal latent variables $\mathbf{z}_{\mathcal{A}_m}$ containing both the modality-specific and shared latent parts.

\paragraph{Decoder:} For each modality, the latent variables $\mathbf{z}_{\mathcal{A}_m}$ are passed through modality-specific decoders to reconstruct the observed variables $\hat{\mathsf{x}}^{(m)}$. The reconstruction loss is then minimized using a quadratic loss in Eq. \ref{rec}, between the true and reconstructed observed variables.
\begin{equation}
\label{rec}
    \mathcal{L}_{rec}=\textbf{$L_2$}(\mathsf{x},\hat{\mathsf{x}})=\underset{m=1}{\overset{M}{\Sigma}}\|\mathsf{x}^{(m)}-\hat{\mathsf{x}}^{(m)}\|_2
\end{equation}

\subsection{Learning the shared latent structure} We leverage a Wasserstein-distance based pluggable differentiable module to capture the partially shared structure from the concatenated latent variables $\displaystyle \mathbf{z}_\text{cat}=\underset{i=1}{\overset{M}{\cup}} \mathbf{z}_{\mathcal{A}_i}$. The latter identifies the shared elements across the different subsets of modalities by learning a permutation matrix $\mathbf{P}$ over $\mathbf{z}_\text{cat}$ defined as follows: each latent variable is mapped to itself when it is modality-specific, and to its corresponding shared component otherwise. 

\noindent{\bf Example 4. }Suppose we have the shared pattern configuration in the left side of Fig. \ref{datasetup} (blue and green part). By duplicating the modality shared variables, one can assume that the latent structure can be written as $\mathbf{z}_\text{cat}=(\mathbf{z}_1,\mathbf{z}_2,\mathbf{z}_7^{(1)},\mathbf{z}_8^{(1)},\mathbf{z}_7^{(2)},\mathbf{z}_8^{(2)},\mathbf{z}_3)$, where $\mathbf{z}_j^{(i)}$ is the duplicate of the shared latent variable $\mathbf{z}_j$ in the modality $i$. The permutation matrix $\mathbf{P}$ is expressed as:\\
$
\mathbf{P} = 
\arrayrulecolor{red}
\begin{blockarray}{ccccc!{\color{red}\vline} ccc}
 & \multicolumn{4}{c}{\text{mod 1}} & \multicolumn{3}{c}{\text{mod 2}}\\
\begin{block}{c[cccc!{\color{red}\vline} ccc]}
\multirow{4}{*}{\text{mod 1}} & \textcolor{red}{1} &\! 0 &\! 0 &\! 0 &\! 0 &\! 0 &\! 0 \\
 & 0 &\! \textcolor{red}{1} &\! 0 &\! 0 &\! 0 &\! 0 &\! 0\\
 & 0 &\! 0 &\! 0 &\! 0 &\! \textcolor{red}{1} &\! 0 &\! 0\\
 & 0 &\! 0 &\! 0 &\! 0 &\! 0 &\! \textcolor{red}{1} &\! 0\\
\cmidrule(lr){2-8}
\multirow{3}{*}{\text{mod 2}} & 0 &\! 0 &\! \textcolor{red}{1} &\! 0 &\! 0 &\! 0 &\! 0\\
 & 0 &\! 0 &\! 0 &\! \textcolor{red}{1} &\! 0 &\! 0 &\! 0 \\
 & 0 &\! 0 &\! 0 &\! 0 &\! 0 &\! 0 &\! \textcolor{red}{1} \\
\end{block}
\end{blockarray}
$\\
The permutation matrix $\mathbf{P}$, hence reveals the partially shared latent elements $\mathbf{z}_4$ and $\mathbf{z}_5$. Non null elements in blocks $[\mathbf{P}]_{(mod 1),(mod2)}$ and  $[\mathbf{P}]_{(mod 1),(mod2)}$ indicate the shared elements between $mod1$ and $mod2$.\\
\textbf{Modeling the permutation learning problem.} In a general case, the permutation matrix $\mathbf{P}$ is obtained through solving the Integer Linear Programming (ILP) problem in Eq. \ref{ilp}. We minimize the trace to find the permutation that minimizes the total transport cost.
\begin{equation}
\label{ilp}
\begin{aligned}
&\underset{\mathbf{P}}{argmin} \quad tr((\mathbf{D}+\epsilon_{epoch}\mathbf{I})\mathbf{P^{T}})\\
& s.t \quad \quad \begin{aligned}
& \mathbf{P}_{ij} \in \{0,1\} \quad \forall\hspace{0.1cm} i,j \in [|1,L|] \\
&  \mathbf{P} \mathbf{1_L}=\mathbf{1_L} \\
&  \mathbf{P^{T}} \mathbf{1_L}=\mathbf{1_L}\\
&  \mathbf{B_{\mathcal{A}_i}^{T}PB_{\mathcal{A}_j}} \leq k \quad \forall\hspace{0.1cm} i,j \in [|1,M|] \hspace{0.1cm} , \hspace{0.1cm} i \neq j
\end{aligned}\\
\end{aligned}   
\end{equation}
where {$[\mathbf{B_{\mathcal{A}_i}}]_k=\begin{cases}
1 \quad if \quad k \in \mathcal{A}_i\\
0  \quad o.w
\end{cases}$} $\epsilon_{epoch}$ is a selection parameter and $\displaystyle L=\underset{i=1}{\overset{M}{\Sigma}} dim(\mathcal{A}_i)$ $[\mathbf{D}]_{ij}=\|(\mathbf{z}_{cat})_i-(\mathbf{z}_{cat})_j\|_2$ is a chosen distance between the $i^{th}$ and $j^{th}$ features in $\mathbf{z}_{cat}$. The first and second constraints in the ILP problem ensure $\mathbf{P}$ is a permutation matrix, whereas the last one fixes the number of latent variables shared by different modalities. The problem in Eq. \ref{ilp} corresponds to an optimal transport problem whose objective is to find the optimal mapping between the latent variables of different modalities $\mathbf{z}_{\mathcal{A}_m}$ and $\mathbf{z}_{\mathcal{A}_n}$. More specifically, it aims to recover the optimal mapping that minimizes the Wasserstein Distance between variables from different modalities, where our associated transport cost matrix:\begin{equation} cost(\mathbf{z}_i,\mathbf{z}_j)=\begin{cases}
\epsilon_{epoch} \quad if \quad i=j\\
\|\mathbf{z}_i-\mathbf{z}_j\|_2  \quad o.w
\end{cases}
\end{equation}
It is noteworthy that $\epsilon_{epoch}$ decreases with the epochs to ensure tighter selection of the shared latent variables.\\
{\bf Solving the problem.} To integrate the ILP problem in our model, we need to relax the binary constraint and solve it with the gradient descent algorithm as in \citep{partiallyaligned}. Note that we need it to be differentiable for backpropagating the gradients. To enforce the above constraints after each gradient update, we use Dykstra's cyclic algorithm to iteratively project the resulting matrix onto the four convex constraint sets, defined in Eq. \ref{projection}.
\begin{equation}
\label{projection}
\begin{aligned}
    & \mathbf{\psi_1(P)}=\mathbf{ReLU(P)}\\
    & \mathbf{\psi_2(P)}=\mathbf{P-\frac{1}{n}(P1_L-1_L)1_L^{T}}\\
    & \mathbf{\psi_3(P)}=\mathbf{P-\frac{1}{L}1_L(1_L^{T}P-1_L^{T})}\\
    & \mathbf{\psi_4^{i,j}(P)}=\mathbf{P-\frac{1}{|\mathcal{A}_i| \cdot |\mathcal{A}_j|}(\mathbf{B_{\mathcal{A}_j}^{T}PB_{\mathcal{A}_i}}-k)B_{\mathcal{A}_i}B_{\mathcal{A}_j}^T}\\
\end{aligned}
\end{equation}
We then inject the resulting permutation matrix $\mathbf{P}$ into the alignment loss $\mathcal{L}_{alg}$ in Eq. \ref{alg}, to pull closer the shared latent variables between the different subsets of modalities.
\begin{equation}
\label{alg}
    \mathcal{L}_{alg}=\|\mathbf{z}_{cat}- \mathbf{z}_{cat}\cdot \mathbf{P}\|_2
\end{equation}
\subsection{Learning the causal structure}
As in \citet{biome}, we estimate the exogenous variables with a Masked Autoregressive Normalizing Flow. The latter receives as input the concatenated causal latent variables, along with a learnable adjacency matrix $J_{\hat{g}}$ and learns the causal mechanisms behind the latent variables. The adjacency matrix encodes the causal generating process, where non null positions $[J_{\hat{g}}]_{ij}$ indicate that the variable $i$ is a parent of the variable $j$. We enforce the learned matrix to be sparse and acyclic by minimizing the following equations:
\begin{equation}
\label{spr}
    \mathcal{L}_{spr}=\|J_{\hat{g}}\|_1
\end{equation}
\begin{equation}
\label{acyc}
    \mathcal{L}_{acy}=tr((\mathbf{I}_L + cst \ (J_{\hat{g}} \circ J_{\hat{g}}))^L) - L 
\end{equation} 
where $cst$ is a constant. The sparsity ensures the component-wise identifiability of the modality-specific and shared causal latent representations as in Th. \ref{componentident}. Acyclicity guarantees that the learned causal graph is a valid DAG.

\subsection{Training objective}
Our model is trained by minimizing the following loss $\mathcal{L}$:
\begin{equation}
    \label{loss}
    \mathcal{L}=\alpha_{alg}\mathcal{L}_{alg}+\alpha_{rec}\mathcal{L}_{rec}+\alpha_{spr}\mathcal{L}_{spr}+\alpha_{acy}\mathcal{L}_{acy}
\end{equation}

\section{Experiments}
To evaluate the efficiency of our approach, we compare it against several baseline methods, including CausalVAE \citep{causalvae}, Multimodal Contrastive Learning (MCL) \citep{mcl}, the Multimodal Biomedical Model \citep{biome} and the Multi-View With Partial Observability framework \citep{multiview} across a diverse set of datasets that encompass realistic synthetic ones. We evaluate the performance of our model through 3 common metrics. \textbf{(a) \textbf{MCC (Mean Correlation Coefficient):}} Measures the component-wise correlation between the estimated and ground-truth latent variables up to permutations. High MCC values close to $1$ reflect near-perfect recovery of the underlying latent structure.\textbf{(b) R² (Coefficient of Determination):} We predict the true latent variables from the estimated ones and report the coefficient of variation. The latter hence quantifies the proportion of variance in the true variables that can be explained by the estimated ones. Higher scores are better. \textbf{(c) EnSHD (Enhanced version of the Structural Hamming Distance):} assesses the similarity between the true and estimated causal graphs. However, since identifiability is established only up to permutation, a direct comparison is not appropriate. We therefore first align the estimated graph according to the permutation that maximizes the MCC. Lower \textbf{EnSHD} is better. 
\subsection{Multimodal Numerical Datasets}
We generate $3$ multimodal numerical datasets, each modality being related to measurements in a different domain. In each case, we have distinct sharing pattern and causal structures to test our framework under different scenarios. We follow the generation process presented in Eq. \ref{generationproc}, where modality-specific functions are implemented by random MLP with LeakyReLU activation functions. Our datasets, respectively, include $2$, $3$ and $4$ modalities. Their detailed presentation is provided in Appendix B.
\begin{table}[ht]
\centering
\caption{Mean$\pm$ standard deviation results of R².}
\label{r2_num}
\arrayrulecolor{black}
\begin{tabular}{p{1.5cm}p{1.8cm}p{1.8cm}p{1.8cm}}
\hline
\textbf{Models} & \textbf{2 Modalities} & \textbf{3 Modalities} & \textbf{4 Modalities}  \\
\hline
\textbf{CausalVAE} & $0.74 \pm 0.09$ & $0.64 \pm 0.04$ & $0.56 \pm 0.14$  \\
\hline
\textbf{MCL} & $0.29 \pm 0.03$ & - & -   \\
\hline
\textbf{MultiBio*} & $0.95 \pm 0.002$ & $0.84 \pm 0.05$ & $0.85 \pm 0.03$  \\
\hline
\textbf{MultiView} & $0.88 \pm 0.16$ & $0.80 \pm 0.06$ & $0.80 \pm 0.07$  \\
\hline
\textbf{Ours} & $\mathbf{0.96 \pm 0.001}$ & $\mathbf{0.88 \pm 0.003}$ & $\mathbf{0.88 \pm 0.02}$  \\
\hline
\end{tabular}
\end{table}
\begin{table}[ht]
\centering
\caption{Mean$\pm$ standard deviation results of the MCC.}
\label{mcc_num}
\arrayrulecolor{black}
\begin{tabular}{p{1.5cm}p{1.8cm}p{1.8cm}p{1.8cm}}
\hline
\textbf{Models} & \textbf{2 Modalities} & \textbf{3 Modalities} & \textbf{4 Modalities}  \\
\hline
\textbf{CausalVAE} & $0.45 \pm 0.01$ & $0.37 \pm 0.01$ & $0.30 \pm 0.01$  \\
\hline
\textbf{MCL} & $0.20 \pm 0.02$ & - & -   \\
\hline
\textbf{MultiBio*} & $0.70 \pm 0.1$ & $0.67 \pm 0.01$ & $0.68 \pm 0.02$  \\
\hline
\textbf{MultiView} & $\mathbf{0.78 \pm 0.2}$ & $0.62 \pm 0.03$ & $\mathbf{0.70 \pm 0.05}$  \\
\hline
\textbf{Ours} & $0.76 \pm 0.02$ & $\mathbf{0.63 \pm 0.03}$ & $0.69 \pm 0.04$  \\
\hline
\end{tabular}
\end{table}
\begin{table}[ht]
\centering
\caption{The EnSHD for the $3$ settings. For all settings the SHD doesn't exceed $8\%$ of the total edges.}
\label{mcc_num}
\arrayrulecolor{black}
\begin{tabular}{p{1.5cm}p{1.8cm}p{1.8cm}p{1.8cm}}
\hline
\textbf{Models} & \textbf{2 Modalities} & \textbf{3 Modalities} & \textbf{4 Modalities}  \\
\hline
\textbf{EnSHD} & $3$ & $7$ & $11$  \\
\hline
\end{tabular}
\end{table}
Tab. \ref{r2_num} and \ref{mcc_num}, respectively, show the R² and MCC coefficients between the true and estimated latent variables. It is noteworthy that the MCL \citep{mcl} framework only accounts for two modalities. We can observe that our proposed model outperforms other SOTA baselines in terms of $R^2$ among all the compared datasets. Hence, it proves that our estimated latent variables under the component-wise identifiability results better explain the ground-true ones, compared to other approaches that only establish block-wise identifiability results \citep{mcl,multiview}, are not multimodal \citep{causalvae} or do not consider overlapping between the latent variables \citep{biome}. We also report better MCC results than the CausalVAE, MCL and MultiBio models and competitive ones with the MultiView. However, the latter is based on contrastive learning and needs carefully constructed negative and positive pairs. 
We also performed an ablation study in the Appendix C to evaluate the effect of causal structural sparsity on the performances of our model. The latter shows our performances remain competitive even when the sparsity constraint is violated and the best performances correspond to sparse scenarios.
\subsection{Synthetic Multimodal Datasets}
\paragraph{Multimodal3DIdent:} We experiment our proposed framework on a more realistic dataset MultiModal3DIdent \citep{mcl} that includes pairs of images along their text annotations. The images are generated from synthetic scenes consisting of a colored object placed in front of a colored background and illuminated by a spotlight of distinct color. More details about the dataset are provided in Appendix B. 
\begin{table}[ht]
\centering
\caption{Mean$\pm$ standard deviation of the MCC and R² on the MultiModal3DIdent dataset.}
\label{mp3dident}
\arrayrulecolor{black}
\begin{tabular}{p{1.5cm}p{1.8cm}p{1.8cm}p{1.8cm}}
\hline
\textbf{Models} & \textbf{MCC} & \textbf{R²}   \\
\hline
\textbf{CausalVAE} & $0.50 \pm 0.01$ & $0.35 \pm 0.1$   \\
\hline
\textbf{MCL} & $0.64 \pm 0.04$ & $0.33 \pm 0.02$    \\
\hline
\textbf{MultiBio*} & $0.78 \pm 0.02$ & $0.44 \pm 0.01$   \\
\hline
\textbf{MultiView} & $\mathbf{0.82 \pm 0.03}$ & $0.46 \pm 0.04$\\
\hline
\textbf{Ours} & $0.81 \pm 0.1$ & $\mathbf{0.50 \pm 0.01}$ \\
\hline
\end{tabular}
\end{table}
Our conclusions from Tab. \ref{r2_num} and \ref{mcc_num} also hold in this realistic setting. Tab. \ref{mp3dident} shows that our approach outperforms the SOTA baselines in terms of $R^2$ for the Multimodal3DIdent dataset. These results indicate that we learn more informative representations. 
\paragraph{Gene Regulatory Network} We evaluate our model on the synthetic single-cell gene expression data generated by SERGIO \citep{sergio}. More details are provided in Appendix B. Our approach here achieves the best $MCC$ scores compared to existing SOTA methods, confirming its applicability to real datasets.
\begin{table}[ht]
\centering
\caption{Mean$\pm$ standard deviation results of the MCC on the Gene Regulatory Network dataset.}
\label{mp3dident}
\arrayrulecolor{black}
\begin{tabular}{p{1.5cm}p{1.8cm}p{1.8cm}p{1.8cm}}
\hline
\textbf{Models} & \textbf{MCC} \\
\hline
\textbf{CausalVAE} & $0.45 \pm 0.03$   \\
\hline
\textbf{MultiBio*} & $0.97 \pm 0.02$  \\
\hline
\textbf{MultiView} & $0.80 \pm 0.03$ \\
\hline
\textbf{Ours} & $\mathbf{0.98 \pm 0.01}$  \\
\hline
\end{tabular}
\end{table}
\section{Conclusion}
This work proposes new component-wise identifiability guarantees for multimodal CRL with partially shared latent structure. More specifically, it shows that under causal structural sparsity constraints, both the shared and group specific latent variables are identified up to component-wise bijections. Although our assumptions restrict the inter-modal causal relationships to some extent, our framework supports the undercomplete setting, which is more flexible. We also introduce a Wasserstein-based module to uncover the partially shared latent structure. Extensive experiments confirm the effectiveness of our method and show that it is a promising approach for learning meaningful causal latent representations.
\newpage
\bibliography{uai2026-template}

\end{document}